\pdfoutput=1

\documentclass[11pt]{article}

\usepackage[]{acl}

\usepackage{times}
\usepackage{latexsym}

\usepackage{caption}
\usepackage{subcaption}
\usepackage{mathtools}
\usepackage{amsmath}
\usepackage{bbm}
\usepackage{mathrsfs}
\usepackage{algorithmicx}
\usepackage{algorithm}
\usepackage{algpseudocode}

\usepackage{array}
\usepackage{caption}
\usepackage{multirow}
\usepackage{booktabs}
\usepackage{textcomp}
\usepackage{graphicx}
\usepackage{enumerate,letltxmacro}
\usepackage[utf8]{inputenc}
\usepackage{csquotes}
\usepackage{nameref}
\usepackage{mdframed}

\DeclareMathOperator*{\argmax}{arg\,max}

\usepackage[T1]{fontenc}

\usepackage[utf8]{inputenc}

\usepackage{microtype}

\usepackage{inconsolata}

\title{A Hypothesis-Driven Framework for the Analysis \\ of Self-Rationalising Models}

\author{Marc Braun \\
  Linköping University\\
  University of Stuttgart\\
  Fraunhofer IPA\\
  \texttt{marc.braun@ipa.fraunhofer.de} \\\And
  Jenny Kunz \\
  Linköping University\\
  \texttt{jenny.kunz@liu.se} \\}

\begin{document}
\maketitle
\begin{abstract}
The self-rationalising capabilities of LLMs are appealing because the generated explanations can give insights into the plausibility of the predictions. However, how faithful the explanations are to the predictions is questionable, raising the need to explore the patterns behind them further.
To this end, we propose a hypothesis-driven statistical framework. We use a Bayesian network to implement a hypothesis about how a task (in our example, natural language inference) is solved, and its internal states are translated into natural language with templates. Those explanations are then compared to LLM-generated free-text explanations using automatic and human evaluations. This allows us to judge how similar the LLM's and the Bayesian network's decision processes are. 
We demonstrate the usage of our framework with an example hypothesis and two realisations in Bayesian networks. The resulting models do not exhibit a strong similarity to GPT-3.5. We discuss the implications of this as well as the framework's potential to approximate LLM decisions better in future work.
\end{abstract}

\section{Introduction}

With the increasing capabilities of large language models (LLMs), more and more tasks that were traditionally solved using human experts and statistical models are now aided by LLMs. Understanding how a model produces its output is an essential factor in the human acceptance of machine learning systems \cite{hci:explainability}. However, understanding the connection between input and output in LLMs is not easily possible \cite{adadi-berrada-survey}.

Recent advances in LLMs generating longer coherent text have popularised self-rationalising models, which produce a natural language explanation (NLE) alongside their output \cite{hase-etal-2020-leakage, DBLP:journals/corr/abs-2111-08284}. 
NLEs have numerous benefits over other, non-textual explanations: NLEs are valued more highly by human users \cite{forrest-etal-2018-towards}, they can be applied to a broad range of problems and they can combine external knowledge with the model input. However, even though the NLEs can give insights into how plausible the predictions made by LLMs are, the \emph{faithfulness} of the explanations to the prediction process remains at best uncertain \cite{wiegreffe-etal-2021-measuring, atanasova-etal-2023-faithfulness, Turpin2023LanguageMD}. 

In this work, we propose exploring the patterns behind generated NLEs using a hypothesis-driven framework, with the ultimate goal of deriving a surrogate model. Our framework is centred around a \emph{hypothetical global explanation} (HGE): A hypothesis about how the LLM solves a specific task on a global, structural level. While we start off with an obviously oversimplified hypothesis to introduce and test the framework, we envision that it can be incrementally adapted to more refined hypotheses in the future. The patterns captured by each refinement step can then serve to measure their coverage, or e-recall \cite{goldberg2023two}, in the LLM. 

The core component of our framework is a statistical surrogate model (SSM) that reflects the HGE.
We propose using a Bayesian Network \cite{pearl1988probabilistic} with a manually designed structure as a framework for the SSM, as the Bayesian Network allows us to visualise the independencies among the random variables used in the SSM via a directed acyclic graph (DAG). This allows us to define the variables in the SSM and the relationships among them such that they reflect the HGE. Furthermore, since the structure of the SSM is based on the HGE, each variable of the Bayesian Network is assigned a specific, semantically interpretable meaning. This allows us to generate local NLEs for individual inputs based on the internal state of the Bayesian Network. 
In the last step, we compare both the predicted labels and the NLEs produced by the SSM to those produced by the LLM in order to gain insights about the faithfulness of the HGE.

We demonstrate the usage of this framework with an exemplary HGE for how the behaviour of the LLM GPT-3.5 \cite{lmsfewshotlearners} can be explained when performing English-language natural language inference (NLI). We discuss the challenges when designing, implementing and training the Bayesian Networks based on the HGE and outline the next steps on the way to a surrogate that models LLM predictions more accurately.

\section{Related Work}
\label{sec:related_work}

Self-rationalising models have received increasing attention as the generation abilities of NLP models have improved in recent years. Human-annotated datasets such as e-SNLI \cite{DBLP:journals/corr/abs-1812-01193} for NLI, 
CoS-E \cite{rajani-etal-2019-explain} and ECQA \cite{aggarwal-etal-2021-explanations} for commonsense question answering and ComVE \cite{wang-etal-2020-semeval} for commonsense validation are the basis for much NLE work \cite{wiegreffe:survey}. However, the role of LLMs in the annotation process itself is likely increasing \cite{wiegreffe-etal-2022-reframing}, as it reduces the significant cost of human annotation \cite{belinkov-glass-2019-analysis}. 

A concern, however, is that the generation of the NLEs is as opaque as the prediction process.
To address this, some recent work explores the faithfulness of NLEs: \citet{wiegreffe-etal-2021-measuring} show that prediction and explanation exhibit correlated responses to input noise.
\citet{atanasova-etal-2023-faithfulness} propose analysing NLEs after counterfactual interventions on the input that alter the prediction, and testing the sufficiency of the reasons provided in the NLE. \citet{Turpin2023LanguageMD} show that biased features in the input lead to obviously unfaithful NLEs as those features affect the predictions heavily but are never mentioned in the generated NLEs. 

Similar concerns have been raised for model-agnostic surrogates such as LIME \cite{DBLP:journals/corr/RibeiroSG16} and SHAP \cite{shap}, which are widely used for highlighting input features as a form of explanation. They approximate a complex model by training a simpler, interpretable model on the original model's predictions. 
Whether such surrogates are helpful for understanding complex models is subject to discussion.
\citet{rudin2019stop} makes the case against their usage in high-stake decisions and argues for the deployment of models with an interpretable decision process. 
However, from a more practical perspective, \citet{jacovi-goldberg-2020-towards} point out that faithfulness is not a binary feature but should be seen as a scale. They argue that depending on the use case, it can be more important to have \emph{plausibility} than \emph{faithfulness}. 

\section{Proposed Framework} 
\label{sec:framework}

\begin{figure}[t]
    \centering
    \includegraphics[width=0.48\textwidth]{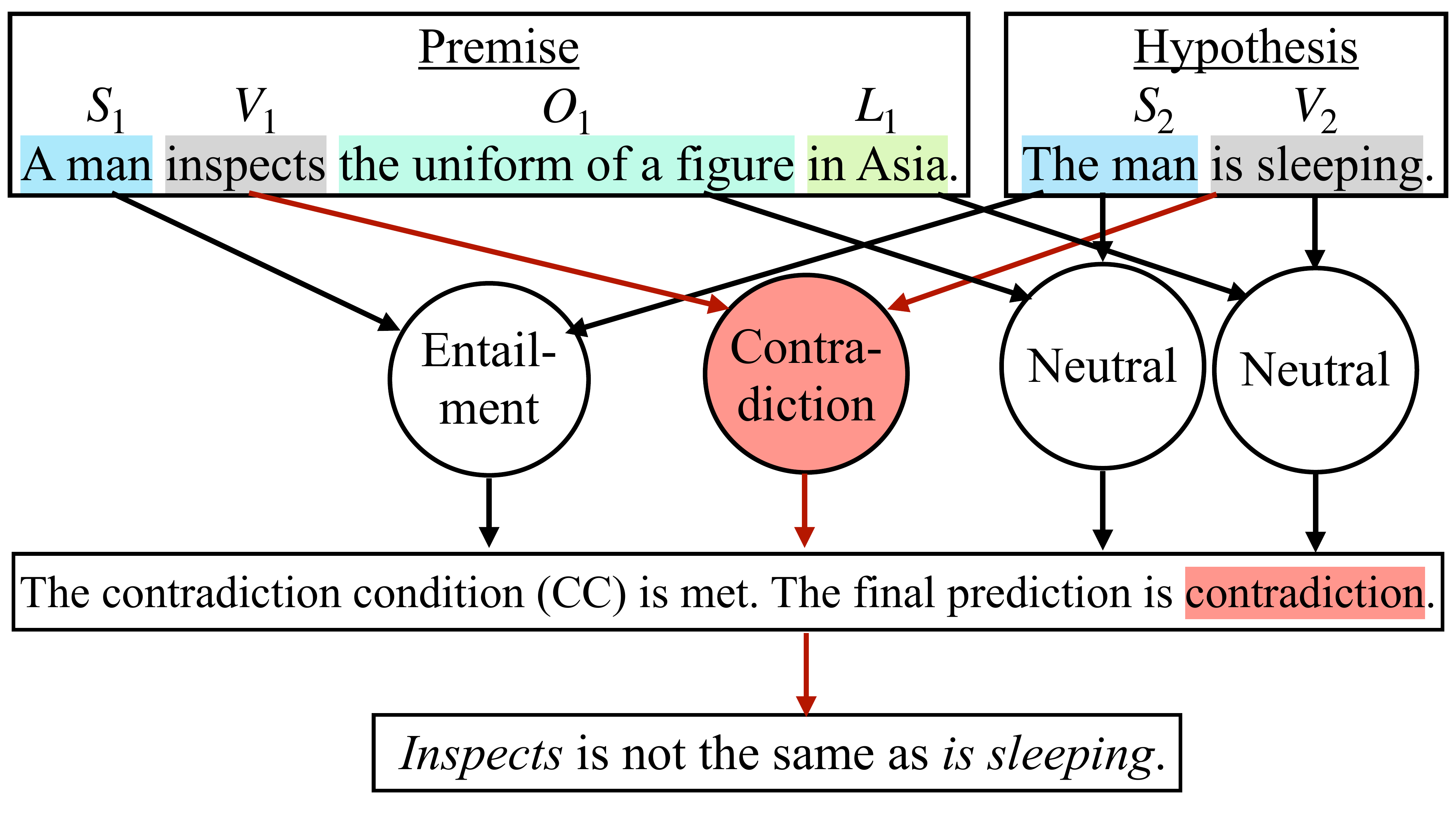}
    \caption{An illustrative (simplified) example for the small SSM. The input $X$ consists of the subphrases of the premise and hypothesis. The circles are the hidden variables $Z$, followed by the final prediction $Y$ (here, \emph{contradiction}) and a template-based NLE (lowest box). }
    \label{fig:SSM_example}
\end{figure}

In this section, we describe our framework\footnote{The implementation used for our experiments is available at \url{https://github.com/Marbr987/Hypothesis_Driven_Analysis_of_Self_Rationalising_Models}.}: How we design the SSM based on our hypothesis (§\ref{sec:ms-2}-§\ref{subsec:bayesian-network-structre}), how we determine and learn the parameters (§\ref{subsec:learn-bn-parameters}) and how we generate NLEs based on the SSM's parameters (§\ref{sec:ms-3}).
We demonstrate that we can successfully control for the assumptions made in our hypothesis, with a model that is intuitive to construct and understand. For a simplified demonstration of how the framework processes an example input, we refer to figure \ref{fig:SSM_example}.

\subsection{Constructing the SSM}
\label{sec:ms-2}

The SSM aims to reflect a hypothesis about a decision-making process that is assumed to explain the behaviour of GPT-3.5 \cite{lmsfewshotlearners}\footnote{We use the \textit{text-davinci-003} version that deviates from the cited paper by being fine-tuned using RLHF \cite{rlhf}, but possibly even in other aspects.} when performing NLI. A hypothetical global explanation (HGE) is defined that aims at providing a hypothesis on how GPT-3.5 produces its labels when performing NLI. The hypothesis is based on an intuitive yet oversimplified assumption about how a fully connected model such as the Transformer network \cite{NIPS2017_3f5ee243} behind GPT-3.5, may solve the NLI task. The HGE is as follows: 

\newpage

\begin{mdframed}[backgroundcolor=gray!20]
\begin{displayquote}
\label{HGE}

\textbf{HGE:} ``When performing NLI, GPT-3.5 compares pairs of subphrases from the premise and hypothesis to each other and classifies each pair into contradiction, entailment, or neutral. Based on this classification, the final prediction is made using deterministic rules.''

\end{displayquote}
\end{mdframed}

\subsection{Extracting Subphrases from Premise and Hypothesis}
\label{subsec:subphrase-extraction}
 
The aim of NLI is to classify the relation of a \emph{premise} and a \emph{hypothesis} into \emph{contradiction}, \emph{entailment}, and \emph{neutral} relation. We use the e-SNLI dataset \cite{DBLP:journals/corr/abs-1812-01193} that besides pairs of premises and hypotheses and the according NLI label also contains human-authored NLEs. An example of a premise and a hypothesis is given in figure \ref{fig:SSM_example}, where the hypothesis ``The man is sleeping'' contradicts the premise ``A man inspects the uniform of a figure in Asia''.

To implement the structure of the \hyperref[HGE]{HGE}, we define the subphrases to be the subject, verb, and object of the sentences as well as location and clothing of the subjects mentioned in the sentences.\footnote{While we chose the categories after a manual inspection of the training data, they are obviously incomplete. More categories would increase the complexity of the Bayesian Network, but ensure better coverage.}
We extract subphrases using the syntactic dependency trees of the sentences as defined by the SpaCy dependency parser \cite{honnibal-johnson-2015-improved}.
Sentences that contain multiple subjects, verbs, or objects were discarded from the dataset. The remaining data contains 30.8\% of the data in the original dataset.
The individual words in each subphrase are transformed into 300-dimensional vectors using Spacy's pre-trained word embeddings \cite{spacy2}. The embedding vectors of the individual words are then added up. If multiple location or clothing subphrases are extracted, the embedding vectors are also added up.

\subsection{Defining the Structure of the SSM}
\label{subsec:bayesian-network-structre}

The random variables (RVs) used in the SSM and the structure of the SSM are deduced from the \hyperref[HGE]{HGE} and defined using a Bayesian Network. A Bayesian Network is a graphical statistical tool to visually represent independencies among RVs. The RVs in a Bayesian Network are represented as nodes in a directed acyclic graph (DAG), where an edge from node $A$ to node $B$ is interpreted as $A$ \textit{causes} $B$.

\paragraph{Defining the input variables $X$.} We represent each subphrase in the premise and hypothesis as a 300-dimensional vector as described in \ref{subsec:subphrase-extraction}.
Let
$S_i, V_i, O_i, L_i$ and $C_i$ be RVs representing the \textit{subject}, \textit{verb}, \textit{object}, \textit{location}, and \textit{clothing} subphrase of sentence $i$,
where sentence 1 is the premise and sentence 2 is the hypothesis.
For notation purposes, we introduce the following random vectors:

\begin{align*}
X_1 &\coloneqq (x_{1,1}, x_{1,2}, x_{1,3}, x_{1,4}, x_{1,5})^T \\
&\coloneqq (S_1, V_1, O_1, L_1, C_1)^T\\
X_2 &\coloneqq (x_{2,1}, x_{2,2}, x_{2,3}, x_{2,4}, x_{2,5})^T \\
&\coloneqq (S_2, V_2, O_2, L_2, C_2)^T,\\
X &\coloneqq (X_1, X_2)^T
\end{align*}

We also define the sets of random vectors:

\begin{align*}
\mathcal{X}_1 &\coloneqq \{S_1, V_1, O_1, L_1, C_1\},\\
\mathcal{X}_2 &\coloneqq \{S_2, V_2, O_2, L_2, C_2\},\\
\mathcal{X} &\coloneqq \mathcal{X}_1 \cup \mathcal{X}_2
\end{align*}

\begin{figure}[ht]
    \centering
    \includegraphics[width=0.48\textwidth]{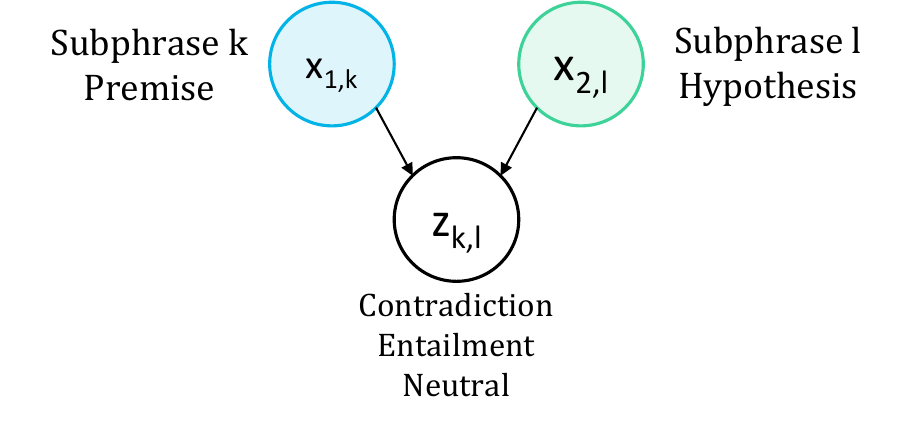}
    \caption{Relationship of any $z_{k,l} \in \mathcal{Z}$ to its parents}
    \label{fig:z-parents}
\end{figure}

\paragraph{Introducing hidden variables $Z$} In order to model the assumption in the \hyperref[HGE]{HGE} that the pairs of subphrases are classified independently, a set of unobserved discrete RVs $\mathcal{Z}$ is introduced. As shown in Figure \ref{fig:z-parents}, we define that each element $z_{k,l} \in \mathcal{Z}$ is caused by $x_{1,k}$ and $x_{2,l}$ where $x_{1,k}$ is the $k$-th element in $X_1$ and $x_{2,l}$ is the $l$-th element in $X_2$. Let $Z$ be the random vectors with entries equal to the elements in $\mathcal{Z}$. According to the \hyperref[HGE]{HGE}, each $z_{k,l}$ is a discrete RV with possible realisations \emph{contradiction}, \emph{entailment}, or \emph{neutral}. In other words, each hidden RV $z_{k,l}$ models the relation of subphrase $x_{1,k}$ from the premise to the subphrase $x_{2,l}$ from the hypothesis.

Depending on which combinations of $k$ and $l$ are included in $\mathcal{Z}$, the structure of the Bayesian Network changes and different interpretations of the \hyperref[HGE]{HGE} can be modelled.
We compare two different structures: A large SSM containing all possible connections, and a small SSM that only contains the connections that we hypothesise are relevant for the final prediction. 

\begin{figure}[t]
    \centering
    \includegraphics[width=0.48\textwidth]{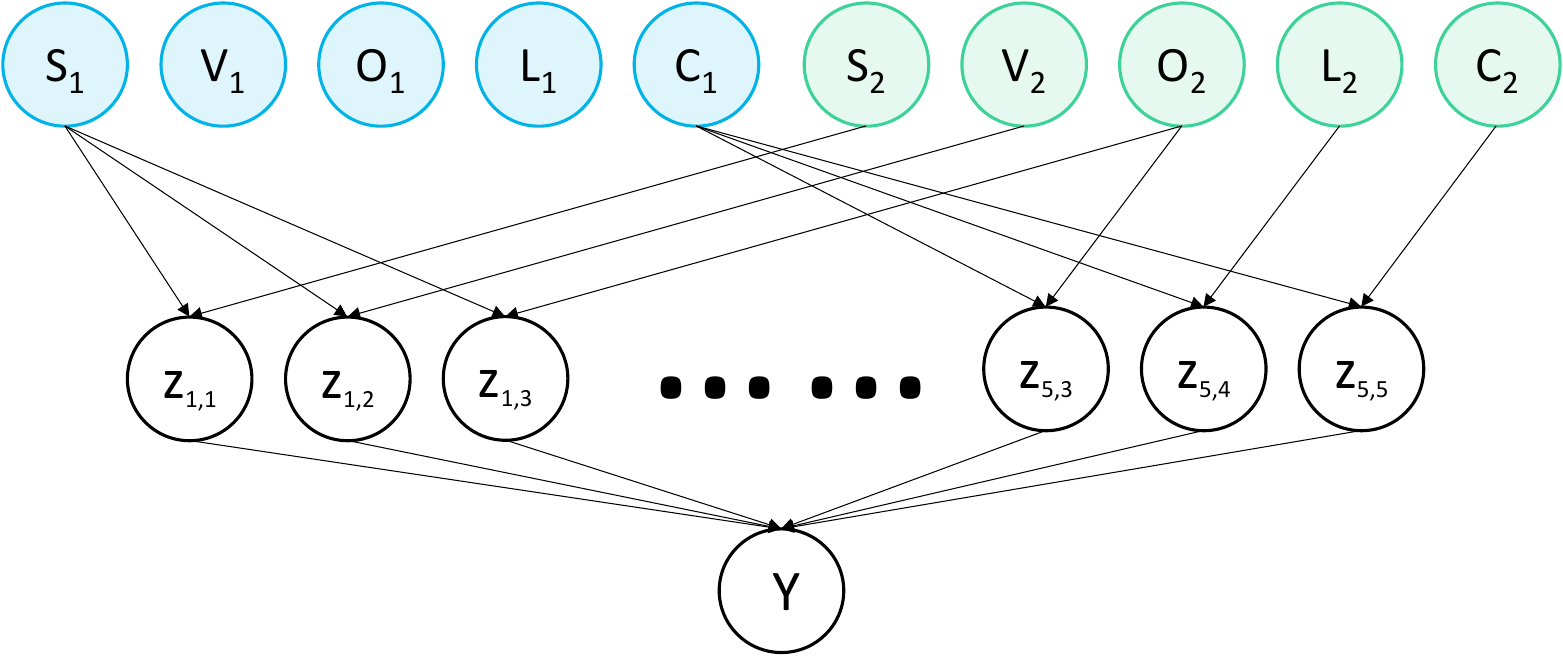}
    \caption{Structure of the $\text{SSM}_{large}$ expressed as a Bayesian Network}
    \label{fig:BN-large}
\end{figure}

\paragraph{Defining $Z$ for the large SSM} For the first structure (visualised in figure \ref{fig:BN-large}), $\mathcal{Z}$ contains all possible combinations of $k$ and $l$, following the assumption that any subphrase from the premise can contradict, entail, or be neutral towards every subphrase in the hypothesis. Let this structure be called \textit{large SSM}. In mathematical terms, we define
$$\mathcal{Z}_{large} \coloneqq \{z_{k,l} | k,l \in \{1,2,3,4,5\} \}$$

\paragraph{Defining $Z$ for the small SSM} For the second structure, we remove all random variables in $\mathcal{Z}$ that we assume to be in the neutral state most of the time and therefore do not contain much relevant information for the prediction. These uninformative $z$ are defined to be the ones that model the \textit{inter}-relationships between the subject, verb, and object subphrases. For example, the subject of the premise rarely contradicts or entails the object of the hypothesis. In mathematical terms, we define
$$\mathcal{Z}_{small} \coloneqq \mathcal{Z}_{large} \setminus \{z_{k,l} | k,l \in \{1,2,3\} \land k \neq l \}$$

\paragraph{Defining the output $Y$} According to the \hyperref[HGE]{HGE}, we derive the final prediction from the subphrase classification using deterministic rules. The results from the subphrase classification are the values of $Z$. Consequently, the final prediction $Y$ is made by defining directed edges from all variables in $\mathcal{Z}$ to $Y$. $Y$ is the discrete RV representing the overall class of the NLI task and therefore has possible states \emph{contradiction}, \emph{entailment}, or \emph{neutral}.

\subsection{Determining the Parameters of the SSM}
\label{subsec:learn-bn-parameters}

The aim of the SSM is to make predictions on $Y|X$, i.e.\ the final class given the subphrases, and to generate NLEs for why the prediction was made. The probability of $Y|X$ can be calculated as follows.

\begin{subequations} \label{eq:y-x}
\begin{align}
P(Y|X)  &= \sum_Z P(Y,Z|X) \label{eq:y-x:l1}\\
        &= \sum_Z P(Y|Z,X) \cdot P(Z|X) \label{eq:y-x:l2}\\
        &= \sum_Z P(Y|Z) \cdot P(Z|X) \label{eq:y-x:l3}
\end{align}
\end{subequations}

The summation is performed over all possible states of the vector $Z$ and the equality in \ref{eq:y-x:l3} holds because of the independence of $Y$ and $X$ given $Z$ which can be verified using the Bayesian Network structure. 
Given Equation \ref{eq:y-x}, we model the distributions of $Y|Z$ and $Z|X$ to make inference on $Y|X$.

\subsubsection{Defining the Deterministic Distribution of \texorpdfstring{$Y|Z$}{Y|Z}}

Following the \hyperref[HGE]{HGE}, the distribution of $Y|Z$ can be modelled with a set of rules we define using propositional logic. Those rules are:

\paragraph{Contradiction Condition} The RV $Z$ models the relationship between the subphrases $X$. If any of the variables in $Z$ has the value \emph{contradiction}, i.e.\ if any subphrase in the premise contradicts any subphrase in the hypothesis, we define the final prediction $Y$ to be \emph{contradiction}.

\begin{equation} \label{eq:contr-cond} \tag{CC}
\exists z \in \mathcal{Z}: (z=\text{``contradiction''})
\end{equation}

We call this formula \textit{contradiction condition} \eqref{eq:contr-cond}. If it evaluates to {\fontfamily{qcr}\selectfont True}, then Y is defined to be \emph{contradiction}.

\paragraph{Entailment Condition} If the final prediction $Y$ is \emph{entailment}, every subphrase in the hypothesis (i.e.\ every element in $\mathcal{X}_2$) has to be entailed by at least one subphrase in the premise (i.e.\ by any element in $\mathcal{X}_1$). 
This condition is expressed as follows:

\begin{equation} \label{eq:ent-cond} \tag{EC}
\begin{aligned}
&\forall l \in \{1,2,3,4,5\} \ \exists k \in \{1,2,3,4,5\}: \\
&\hspace{2.5cm} (z_{k,l}=\text{``entailment''})
\end{aligned}
\end{equation}

We call this formula \emph{entailment condition} or \eqref{eq:ent-cond}. If it evaluates to {\fontfamily{qcr}\selectfont True}, then Y is defined to be \emph{entailment}.

\paragraph{Neutral Condition} If the  \eqref{eq:contr-cond} and  \eqref{eq:ent-cond} are both evaluated as {\fontfamily{qcr}\selectfont False}, then we define the final class to be \emph{neutral}. The logical formula for this condition can be expressed as

\begin{equation} \label{eq:neutr-cond} \tag{NC}
\neg \eqref{eq:contr-cond} \land \neg \eqref{eq:ent-cond}
\end{equation}

and is called \emph{neutral condition} or \eqref{eq:neutr-cond}.

\paragraph{Distribution of $Y|Z$} Given these three conditions, the distribution of $Y|Z$ can be described as

\begin{equation} \label{eq:y-z}
\begin{aligned}
&P(Y=\text{``contradiction''}|Z)  = 1 \text{ if } \eqref{eq:contr-cond}\\
&P(Y=\text{``entailment''}|Z)  = 1 \text{ if } \eqref{eq:ent-cond}\\
&P(Y=\text{``neutral''}|Z)  = 1 \text{ if } \eqref{eq:neutr-cond}
\end{aligned}
\end{equation}

\subsubsection{Learning the Parameters \texorpdfstring{$\theta$}{t} of the Stochastic Distribution of \texorpdfstring{$Z|X, \theta$}{Z|X, t}}

To be able to evaluate Equation \ref{eq:y-x}, we must determine the distribution of $Z|X$. Let the parameters of this distribution be $\theta$, i.e.\ the goal is to find $\theta$ that models the distribution of $Z|X, \theta$. Since all random vectors in $\mathcal{Z}$ are independent of each other given $X$, it suffices to model the distribution of each individual random vector $z_{k,l} \in \mathcal{Z}$. In this paper, the distribution of each $z_{k,l}|X, \theta$ is modelled by a feed-forward neural network. From the structure of the Bayesian Network we know that each $z_{k,l}$ only has parents $x_{1,k}$ and $x_{2,l}$ which means that to model $z_{k,l}|X, \theta$, each neural network only needs to take $x_{1,k}$ and $x_{2,l}$ as input.


To optimise the parameters $\theta$, we use the EM algorithm \cite{10.2307/2984875}.
In Appendix \ref{subsec:em-alg} we show that the EM algorithm for the problem at hand amounts to taking samples $\tilde{Z}$ from the distribution of $Z|X, Y, \theta^{(t)}$ in the E-step and maximising the log-likelihood of the parameter $\theta$ with respect to the sampled values $\tilde{Z}$ in the M-step.
Since each distribution $z_{k,l} | X, \theta_{k,l}$ is modelled using a neural network, the M-step amounts to training each network on the sampled values $\tilde{z}_{k,l}$ as output and the respective observed values of $X$ as input using the cross-entropy loss function \cite{Boer:2005aa}.

\subsection{Generating NLEs Using the SSM}
\label{sec:ms-3}

In the last step, NLEs are produced using the SSM. By comparing these NLEs with the NLEs produced by GPT-3.5 in \ref{sec:ms-1}, we gain insights about the extent to which the SSM reflects a similar reasoning path as the NLEs produced with GPT-3.5.

By defining the distribution of $Y|Z$ based on a set of rules, we can deduce NLEs from the random vectors in $\mathcal{Z}$. For example, the \eqref{eq:contr-cond} formula states that if any of the random vectors in $\mathcal{Z}$ is classified as a contradiction, then $Y$ is of class contradiction as well. This means that if, for example, the random vector $z_{k,l} \in \mathcal{Z}$ is of class contradiction (and assuming all other elements in $\mathcal{Z}$ are of class other than contradiction), it is reasonable to state that $z_{k,l}$ is responsible for the final prediction.
This example illustrates that, by defining the relationship between $Z$ and $Y$ in a way that is semantically interpretable, explanations for the final prediction can be formulated that follow the predefined semantics of the hidden variables.

Based on the rules used to define the distribution $Y|Z$, the templates that provide reasons for the final prediction $Y$ are defined. The complete templates can be found in Appendix \ref{app:ssm-explanation-templates}.

\section{Comparison to the LLM}
Now that we have constructed the SSM, we need to compare it to GPT-3.5. To that end, we first generate labels and explanations with GPT-3.5 (§\ref{sec:ms-1}), and compare them to the SSM in human (§\ref{subsec:human-evaluation}) and automatic (§\ref{subsec:auto-evaluation}) evaluations. 

\subsection{Generating Labels and Explanations with GPT-3.5}
\label{sec:ms-1}

To compare the outputs of GPT-3.5 to the outputs of SSM, we prompt the GPT-3.5 model to generate entailment labels as well as NLEs that provide reasoning for why the respective label was chosen. 

Performing NLI with GPT-3.5 using few-shot learning has shown to have a performance close to fine-tuned GPT-3.5 \cite{DBLP:journals/corr/abs-2111-08284}. Inspired by the prompts proposed by \citet{DBLP:journals/corr/abs-2111-08284}, we use the following structure.

First, the instruction ``Classify into entailment, neutral, and contradiction and justify the decision.'' is given to the model, followed by six examples\footnote{While the original work by \citet{DBLP:journals/corr/abs-2111-08284} uses 50 examples, we found six to be sufficient for the newer, instruction-tuned GPT models.}, each of which has the following structure:

\begin{minipage}[t]{0.8\textwidth}
    ``Premise: \textit{premise} \\
    Hypothesis: \textit{hypothesis}\\
    Label: \textit{label} \\
    Explanation: \textit{explanation} '' 
\end{minipage}\\

The examples are balanced among the classes, i.e.\ two examples are chosen at random from the training set for each class. The examples are followed by the premise and hypothesis which shall be classified.
Based on this input, GPT-3.5 produces a label and an NLE.

\subsection{Evaluating the Similarity Between the SSM and GPT-3.5 for NLI}
\label{sec:ms-4}

In comparison to surrogate models for input feature highlighting surrogates such as LIME, the proximity of our surrogates to the original model cannot be measured by performance only.
Therefore, we introduce an evaluation setup consisting of a human evaluation and a set of automatic metrics that compare the predictions and NLEs produced by GPT-3.5 to the predictions and NLEs produced by the SSM. This setup allows us to assess the extent to which the \hyperref[HGE]{HGE} is a valid explanation for how GPT-3.5 solves the NLI task.

\subsubsection{Human Evaluation}
\label{subsec:human-evaluation}

For the human evaluation, three raters were instructed to answer the following questions for a random sample of 100 explanations produced for the development set. The full instructions for the raters can be found in Appendix \ref{sec:human-instructions}.

The raters were asked to report which of the subphrases defined in Section \ref{subsec:subphrase-extraction} are mentioned in each NLE produced by GPT-3.5. The reported subphrases can then be compared to the subphrases used in the explanation produced by the SSM. Furthermore, annotators were instructed to indicate whether the GPT-3.5 explanation relates any subphrases from the premise or hypothesis to each other, i.e.\ if it has a structure similar to what is stated in the \hyperref[HGE]{HGE}.

Additionally, the raters were asked to assess whether the GPT-3.5 NLE supports the predicted label and whether the explanation is factually correct.
Lastly, the factual correctness of the NLEs produced by the SSM was assessed by the raters. 
To assess the inter-rater agreement among the human raters, we report Fleiss' Kappa \cite{fleiss1971measuring}.




\subsubsection{Automatic Evaluation}
\label{subsec:auto-evaluation}

In addition to the human evaluation, we use a range of methods to automatically assess the alignment of the SSMs with GPT-3.5.

\paragraph{Similarity Between the Predicted Labels}

To compare the similarity between the predictions made by GPT-3.5 and by the SSMs, we report the accuracy and F1-Score for all predictions, Cohen's kappa \cite{cohen1960} as well as the precision and recall for each class.

\paragraph{Similarity Between the Explanations}

To quantitatively analyse how similar the explanations produced by GPT-3.5 and the SSM are, the cosine similarity between the NLEs, the Jaccard similarity and the BERTScore \cite{DBLP:journals/corr/abs-1904-09675} are calculated.
We report the mean similarity values of all three similarity measures between the NLEs produced by GPT-3.5 and the NLEs produced by the large and small SSM, respectively.
To provide additional context for the similarity measures, the same similarity measures are also calculated between the gold standard NLEs and the NLEs produced by GPT-3.5.

\section{Results}
\label{sec:results}

In this section, we report the experimental results using the evaluation procedures and measures introduced in Section \ref{sec:ms-4}, assessing how similar our surrogate models are to GPT-3.5. For illustratory purposes, we will provide example explanations generated by the SSMs in §\ref{sec:res_example}, followed by the results of the human (§\ref{res:humeval}) and the automatic evaluations (§\ref{res:quanteval}).

\subsection{Example Outputs of the SSMs}
\label{sec:res_example}

The NLEs produced by the SSM do indeed have a structure that follows the \hyperref[HGE]{HGE}. For the premise ''A young woman sits crosslegged beside her purse on the grass among a crowd of dogs.`` and the hypothesis ''The woman is on the couch with the dogs.``, both SSMs generate ''Grass is not the same as couch.`` The semantics of this example NLE capture the reason why there is a contradiction, namely that the location of the woman is different (and thereby contradictory) in the two statements.

Overly complex explanations are a challenge, particularly for the entailment relation. For the premise ''A man tries to get himself into shape on a treadmill.`` and the hypothesis ''A man exercising.``,  the small SSM's explanation is ''Man is the same as a man and get is the same as exercising and if the location of sentence 1 is treadmill, then the verb of sentence 2 has to be exercising.``, which is not only partially incorrect but also points to trivial information.

In Appendix \ref{app:example-outputs} we provide more examples for how the output of the SSM looked like.

\subsection{Human Evaluation}
\label{res:humeval}

The human evaluation revealed that the average Jaccard similarity between the subphrases used in the NLEs by GPT-3.5 and the SSM is 0.384 for the large SSM and 0.423 for the small SSM. The average Fleiss' Kappa for detecting the subphrases used in the NLEs by GPT-3.5 is 0.640. In the remainder of this section, we report Fleiss' Kappa for the three raters in parentheses following the average rating.
The raters found that 84.2\% (0.316) of explanations of GPT-3.5 follow a structure that, as stated in the \hyperref[HGE]{HGE}, relates different subphrases from premise and hypothesis to each other and that 96.8\% (0.858) of NLEs support the predicted label. The raters marked 86.3\% (0.287) of GPT-3.5's NLEs, 27.4\% (0.553) of the NLEs produced using the small, and 21.1\% (0.522) of the NLEs produced by the large SSM as factually correct.

Overall, the results from the human evaluation indicate comparatively low correctness and low similarity of the SSM explanations, although a large part of the explanations follows a structure that would be possible to model with a Bayesian Network. The small SSM's NLEs are more similar to GPT-3.5's NLEs than the large SSM's.

\subsection{Automatic Evaluation}
\label{res:quanteval}

This section presents the results of the automatic evaluation, where different measures were calculated to assess the similarity between the outputs produced by GPT-3.5 and the SSM.

\subsubsection{Similarity Between Predicted Labels}

All analysed metrics shown in Table \ref{tab:pr-evaluation-labels} reveal that the small SSM tends to predict the label of GPT-3.5 with a higher precision and recall (except for precision of class contradiction and recall of class neutral). The table also shows that the values are much lower than the values of the metrics between the gold standard label and the predictions made by GPT-3.5.

\begin{table}[ht]
\small
  \centering
  \begin{tabular}{lccc}
    \toprule
    \multicolumn{1}{l}{Metric} & \textit{Gold} & $\text{SSM}_{\text{large}}$ & $\text{SSM}_{\text{small}}$ \\
    \midrule
    Contr. Precision & \textit{0.949} & \textbf{0.763} & 0.713 \\
    Contr. Recall & \textit{0.899} & 0.286 & \textbf{0.576} \\
    \midrule
    Entail. Precision & \textit{0.968} & 0.619 & \textbf{0.667} \\
    Entail. Recall & \textit{0.704} & 0.549 & \textbf{0.573} \\
    \midrule
    Neutr. Precision & \textit{0.485} & 0.242 & \textbf{0.299} \\
    Neutr. Recall & \textit{0.887} & \textbf{0.630} & 0.528 \\
    \midrule
    Accuracy & \textit{0.807} & 0.468 &  \textbf{0.556} \\
    Avg. F1-Score & \textit{0.788} & 0.449 & \textbf{0.545} \\
    Cohen's Kappa & \textit{0.709} & 0.219 & \textbf{0.339} \\ \bottomrule
  \end{tabular}
  \caption{Metrics relating the SSMs' to GPT-3.5's predictions. The interval for Cohen's Kappa is between -1 and 1 (with 0 being random and 1 perfect agreement) and for all other metrics from 0 to 1.}
  \label{tab:pr-evaluation-labels}
\end{table}

\subsubsection{Similarity Between NLEs}

The similarity scores between the NLEs produced by GPT-3.5 and the NLEs produced using the SSM can be seen in Table \ref{tab:sim-scores-explanations}.

\begin{table}[ht]
  \small
  \centering
  \begin{tabular}{lccc}\toprule
    Model & Cosine S.\ & Jaccard S.\ & BERTScore\ \\
    \midrule
    \textit{Gold } & \textit{0.808} & \textit{0.277} & \textit{0.604} \\
    $\text{SSM}_{\text{large}}$ & 0.771 & 0.182 & 0.455 \\
    $\text{SSM}_{\text{small}}$ & \textbf{0.779} & \textbf{0.196} & \textbf{0.463} \\
    \bottomrule
  \end{tabular}
  \caption{Similarity Scores in relation to GPT-3.5 NLEs. The interval for all metrics is from 0 to 1.}
  \label{tab:sim-scores-explanations}
\end{table}

It is clear that for all three metrics, the predictions made by GPT-3.5 are closer to the gold standard NLEs than to the NLEs of the SSM. As in the human evaluation, the self-rationalised NLEs by GPT-3.5 are closer to the NLEs by the small SSM compared to the NLEs by the large SSM.

\section{Discussion}
\label{sec:discussion}

This paper set out with the aim of constructing and evaluating a hypothesis-driven surrogate for how GPT-3.5 performs NLI. 
As we have seen in section \ref{sec:framework}, the Bayesian network allows us to incorporate our hypothesis in an intuitive and testable way, making (in-)dependencies between phrases explicit. 
While the framework is convenient and functional, the results in section \ref{sec:results} indicate a low similarity to GPT-3.5. This can have two reasons: First, that the HGE does not resemble the way GPT-3.5 solves the NLI task, or second, that implementation details of the SSM have shortcomings that limit the performance of the model.
In any case, several challenges need to be addressed before an SSM can potentially pass as a surrogate model. We discuss the shortcomings of our models as well as paths forward in §\ref{sec:disc_results} and \ref{sec:disc_unique}, and promising paths for other future research based on our insights from this paper in \ref{sec:disc_future_work}.

\subsection{Results and Similarity } 
\label{sec:disc_results} 
Human raters found the factual correctness of the NLEs produced by the SSMs to be very low compared to GPT-3.5 (§\ref{res:humeval}). The low alignment between the outputs of SSM and GPT-3.5 as observed in §\ref{res:humeval} and §\ref{res:quanteval} suggests that the faithfulness of the HGE is relatively low.

As we kept the setup for this paper straightforward, many simplifying assumptions and tradeoffs have been made, limiting the expressiveness on several ends: The hypothesis is kept simple, the coverage of the preprocessing and the template-based NLE generation are limited, and the models used are not optimised. We expect that work on these fronts can substantially improve the model, especially given that our human raters found that a large share of GPT-3.5's explanations follows a structure that relates subphrases from premise and hypothesis to each other (§\ref{subsec:human-evaluation}), as reflected in the assumption behind the HGE. 
Another indication that development time investments may pay off is a recent work by \citet{stacey2023logical} who propose an interpretable model based on logical rules similar to what is stated to the HGE defined in this paper. The high performance on the NLI task this model achieves suggests that an adapted HGE that reflects their model could have a considerably higher overlap with GPT-3.5 predictions than our current SSMs. 

\subsection{Uniqueness of the surrogate model }
\label{sec:disc_unique}
We found that there was no unique surrogate model for our HGE. The room for interpretation given the HGE formulated in natural language allows us to deduce a large number of structures, two of which we implemented. 
We showed that our different interpretations of the HGE (our small and large SSM) lead to different implementations of the surrogate models. However, to fully account for the ambiguity of the HGE when estimating the faithfulness, all viable SSMs would need to be constructed and compared to the LLM outputs.

The small model that incorporates more inductive biases performed better and produced NLEs more similar to GPT-3's. This may indicate that the large model relied on incorrect cues introduced by less relevant connections. If this applies, it indicates that even the training regime of the SSM is of high importance in order for it to correctly represent the HGE. As we did not employ regularisation strategies in the feed-forward networks that we train in the Bayesian Network, this could be a path forward for potential improvements.

\subsection{Future Work}
\label{sec:disc_future_work}

The SSMs tested in this paper do not show sufficient similarity with the original model yet. Apart from altering the model or hypothesis as discussed in §\ref{sec:disc_results} and §\ref{sec:disc_unique}, we see other directions for future research building on this paper:

\paragraph{ Estimating the uncertainty of the faithfulness }
As previously mentioned, it is typically not possible to deduce one unique SSM from an HGE formulated in natural language. Consequently, there is uncertainty in the faithfulness estimation that is gained by comparing the outputs of the LLM with one single SSM. Future research could investigate how this uncertainty might be estimated.
For example, the ambiguity of the HGE would need to be taken into account when estimating this uncertainty. Research on investigating vagueness and ambiguity of written text samples is already an active field of research \cite{freitas-etal-2015-hard,wang-agichtein-2010-query,bernardy-etal-2018-compositional}. How these metrics can be used as a statistical measure of the uncertainty of an HGE remains an open question.

\paragraph{Automatically deriving surrogate models } We conducted a case study for how an SSM can be constructed for an example HGE when performing English-language NLI. However, our manual design has the consequence that in many cases, different SSMs need to be constructed for new tasks, languages and HGEs.\footnote{For a detailed discussion, we refer to §\ref{sec:limitations}.} Therefore, in contrast to methods like LIME \cite{DBLP:journals/corr/RibeiroSG16} that can be applied directly to any classifier, defining a generalised procedure for how to automatically construct suitable surrogate models is challenging and requires further research.

\section{Conclusion}

This paper suggests a framework for testing a hypothesis about the decision-making process of LLMs using Bayesian networks. We demonstrate how to construct a Bayesian network based on a hypothetical global explanation and how to evaluate the alignment of the LLM with this network. The framework intuitively implements the hypothesis as the Bayesian Network that it is based on can model our assumptions natively, and its random variables can be translated into natural language.

Our surrogates were however not similar to GPT-3.5, which can have two reasons: That the hypothesis itself does not accurately describe GPT-3.5's behaviour and needs to be replaced or complemented by other hypotheses, or that simplifications we made in our implementation affected the performance and thereby also the similarity to GPT-3.5. We assume that we are dealing with a combination of the two reasons, and suggest that working on the identified issues can lead to more accurate surrogate models that can help us understand the behaviour of LLMs better. 

\section{Limitations}
\label{sec:limitations}

In this section, we address further limitations of our work and outline potential paths to overcome them in future work. 

\paragraph{Task and Data }Natural language inference with the e-SNLI dataset is a task that naturally fits into the framework of a Bayesian Network, as most explanations are heavily built on input phrases and implicitly follow template-like structures \cite{DBLP:journals/corr/abs-1812-01193}. Many tasks, particularly such that include facts and commonsense reasoning not explicitly stated in the input, are substantially harder to model with such intuitive and simple structures. 

SNLI \cite{bowman-etal-2015-large}, the base dataset for e-SNLI, has been shown to include various annotation artefacts that models can rely on \cite{gururangan-etal-2018-annotation}. While our focus is on the evaluation of the explanations and not on the performance on the prediction task and the Bayesian Network by design cannot pick up the cues stated in the paper, it cannot be excluded that such artefacts have had an influence on the results of our automatic evaluations.
For future work, it may be worthwhile to also consider alternatives, such as the more diverse MNLI dataset \cite{williams-etal-2018-broad}, and include explicit tests for lexical and syntactic cues, e.g.\ with the HANS dataset \cite{mccoy-etal-2019-right}.

\paragraph{Language} English-language systems cannot always be trivially adapted to other languages, particularly where orthographic system, syntax and morphology differ substantially \cite{munro-manning-2010-subword}. 
As a result, our structure may be a worse fit for other languages with different features. In any case, our rule-based preprocessing would need to be adapted. While MNLI \cite{williams-etal-2018-broad} has been translated into a relatively diverse set of languages \cite{conneau-etal-2018-xnli}, there are currently no human-annotated NLEs for this data set. As we work with few-shot prompts for generating the LLM's explanations, this may however be overcome with relatively little work. 

\paragraph{Human Evaluation} Our human evaluation is conducted by a small set of annotators with a machine learning-related background and similar demographics. This is a common practice in NLP research, but it introduces sampling bias \cite{van-der-lee-etal-2019-best}. While we do not explicitly ask for personal preferences, this may still affect their judgements and thereby the results of our evaluation. A larger set of more diverse annotators, if feasible, is preferable.

\paragraph{Reproducibility} We use GPT-3.5, a closed-source model by OpenAI that we only have API access to. Unfortunately, this limits our experiments' reproducibility, as OpenAI may remove or restrict access to it in the future. At the time of writing, the GPT-3.5 model produced better-quality output for our purposes than its more open competitors. However, in recent months, an increasing number of high-quality LLMs is released to the public, such as various LLaMA \cite{touvron2023llama}-based models such as Alpaca \cite{alpaca} or the Pynthia \cite{biderman2023pythia}-based Dolly model \cite{DatabricksBlog2023DollyV2}, paving the way for more reproducible LLM research.

\section*{Ethics Statement}
As LLMs are trained on human data that can be biased, toxic, and immoral, they frequently produce unethical outputs \cite{liu-etal-2022-aligning, anti-muslim-bias, parrots}. However, we use LLMs solely as an object of examination. This study aims at increasing the transparency and accountability of GPT-3.5, which can be a step in the direction of preventing LLMs from producing unethical outputs.
That said, explainability techniques for models as large as current LLMs are only an approximation. We do not endorse any usage of LLMs for high-stake applications without humans in the loop, even as explainability research progresses.

\section*{Acknowledgements}

We thank the anonymous reviewers for their constructive feedback that helped to refine this work.

\bibliography{anthology,references}

\appendix

\section{Appendix}
\label{sec:appendix}

\subsection{EM-Algorithm}
\label{subsec:em-alg}

\subsubsection*{E-Step}

Given the observed variables $X, Y$, the hidden variable $Z$, and the parameters $\theta$, the E-step of the EM-algorithm can be expressed in the following way:

\begin{equation} \label{eq:q-start}
\small
\begin{aligned}
Q(\theta|\theta^{(t)}) &= E_{Z|X,Y,\theta^{(t)}}\left[\log \mathcal{L}(X,Y,Z|\theta)\right]\\
&= E_{Z|X,Y,\theta^{(t)}}\left[\sum_{i=1}^n \log P(x_{(i)},y_{(i)},Z|\theta)\right]\\
&= \sum_{i=1}^n E_{Z|X,Y,\theta^{(t)}}\left[\log P(x_{(i)},y_{(i)},Z|\theta)\right]\\
&= \sum_{i=1}^n \sum_{Z} P(Z|x_{(i)},y_{(i)}, \theta^{(t)}) \\
& \hspace{2.5cm} \cdot \log P(x_{(i)}, y_{(i)}, Z | \theta) \\
\end{aligned}
\end{equation}

In Equation \ref{eq:q-start}, $x_{(i)}, y_{(i)}$ is the $i$-th observed realisation of the random vector $X$ and $Y$ respectively and $n$ is the total number of observations. The summation over $Z$ is performed over all possible states of the discrete random vector $Z$. As all valid probability distributions must sum up to one over their domain, we know that $\sum_{Z} P(Z|x_{(i)}, y_{(i)}, \theta^{(t)}) = 1$. This makes the expression $\sum_{Z} P(Z|x_{(i)}, y_{(i)}, \theta^{(t)}) \cdot \log P(x_{(i)}, y_{(i)}, Z | \theta)$ a weighted average for a given $i$ where the weight is $P(Z|x_{(i)}, y_{(i)}, \theta^{(t)})$.

This means that we can approximate this weighted average by calculating the mean of $\log P(x_{(i)}, y_{(i)}, \tilde{Z} | \theta)$ where $\tilde{Z}$ are $Z$ samples from the distribution given by the weights (i.e.\ $P(Z|x_{(i)}, y_{(i)}, \theta^{(t)})$). Consequently we can approximate $Q(\theta | \theta^{(t)})$ as
\begin{equation} \label{eq:q-appr}
\small
\begin{aligned}
Q(\theta|\theta^{(t)}) &\approx \sum_{i=1}^{n} \frac{1}{s} \sum_{\tilde{Z}_{(i)}} \log P(x_{(i)}, y_{(i)}, \tilde{Z}_{(i)} | \theta)
\end{aligned}
\end{equation}
where $\tilde{Z}_{(i)}$ are $s$ sampled values from the distribution of $Z|x_{(i)}, y_{(i)}, \theta^{(t)}$. This distribution can be expressed as

\begin{equation}
\small
\begin{aligned}
P(Z|x_{(i)}, y_{(i)}, \theta^{(t)}) &= \frac{P(Z,y_{(i)} | x_{(i)}, \theta^{(t)})}{P(y_{(i)} | x_{(i)}, \theta^{(t)})} \\
&\propto P(Z, y_{(i)} |  x_{(i)}, \theta^{(t)}) \\
&= P(y_{(i)} | Z, x_{(i)}, \theta^{(t)}) \\
& \hspace{2.5cm} \cdot P(Z | x_{(i)}, \theta^{(t)}) \\
&\text{Independence of } Y \text{ and } X, \theta \text{ given } Z\\
&= P(y_{(i)} | Z) \cdot P(Z | x_{(i)}, \theta^{(t)})
\end{aligned}
\end{equation}

The distribution of $Y|Z$ was defined in Equation \ref{eq:y-z}. The distribution of $Z | X, \theta^{(t)}$ is known assuming the current parameter estimate $\theta^{(t)}$ is the true parameter of the distribution. Furthermore, we know that $P(Y|Z)$ is either of value 0 or 1. We can therefore produce a sample $\tilde{z}$ from $Z | X, Y, \theta^{(t)}$ by sampling from $Z | X, \theta^{(t)}$ and rejecting the sample if $P(Y|\tilde{z}) = 0$. If the sample is rejected, we repeat the sampling process until a sample is accepted.

Let $s$ be the number of samples we generated for each $i$ from $Z | x_{(i)}, y_{(i)}, \theta^{(t)}$, $n$ the number of observations in the dataset of $x$ and $y$, and let $\tilde{Z}$ be the collection of all $n \cdot s$ samples. Let $\tilde{z}_j$ be the j-th sample in $\tilde{Z}$. We then produce $s$ duplicates of each $x_{(i)}, y_{(i)}$ and define $x_{(j)}, y_{(j)}$ to be those datapoints $x, y$ that were used to produce the sample $\tilde{z}_j$. Based on that definition, we can write Equation \ref{eq:q-appr} as
\begin{equation} \label{eq:final-q}
\small
\begin{aligned}
Q(\theta|\theta^{(t)}) &\approx \frac{1}{s}  \sum_{j=1}^{n \cdot s} \log P(x_{(j)}, y_{(j)}, \tilde{z}_{(j)} | \theta) \\
&= \frac{1}{s}  \sum_{j=1}^{n \cdot s} \log \left[P(y_{(j)}, \tilde{z}_{(j)} | x_{(j)}, \theta) \cdot P(x_{(j)} | \theta)\right] \\
&\text{Independence of } X \text{ and } \theta \\
&= \frac{1}{s}  \sum_{j=1}^{n \cdot s} \log \left[P(y_{(j)}, \tilde{z}_{(j)} | x_{(j)}, \theta) \cdot P(x_{(j)})\right] \\
&= \frac{1}{s}  \sum_{j=1}^{n \cdot s} \log \big[P(y_{(j)} | \tilde{z}_{(j)}, x_{(j)}, \theta) \\
& \hspace{2.5cm} \cdot P(\tilde{z}_{(j)} | x_{(j)}, \theta) \cdot P(x_{(j)})\big] \\
&\text{Independence of } Y \text{ and } X, \theta \text{ given } Z\\
&= \frac{1}{s}  \sum_{j=1}^{n \cdot s} \log \big[P(y_{(j)} | \tilde{z}_{(j)}) \\
& \hspace{2.5cm} \cdot P(\tilde{z}_{(j)} | x_{(j)}, \theta) \cdot P(x_{(j)})\big] \\
& P(y_{(j)} | \tilde{z}_{(j)}) = 1 \text{ for the sampled } \tilde{Z}, \text{ because} \\
&\text{otherwise the sample got rejected}\\
&= \frac{1}{s}  \sum_{j=1}^{n \cdot s} \log \left[P(\tilde{z}_{(j)} | x_{(j)}, \theta) \cdot P(x_{(j)})\right] \\
\end{aligned}
\end{equation}

\subsubsection*{M-Step}

In the M-step, the estimation for $\theta$ is updated by setting $\theta^{(t+1)}$ to the value that maximises $Q(\theta|\theta^{(t)})$.

\begin{equation} \label{eq:theta-update}
\small
\begin{aligned}
\theta^{(t+1)} &= \argmax_{\theta} Q(\theta|\theta^{(t)}) \\
&\text{With Equation \ref{eq:final-q}} \\
&\approx \argmax_{\theta} \frac{1}{s}  \sum_{j=1}^{n \cdot s} \log \left[P(\tilde{z}_{(j)} | x_{(j)}, \theta) \cdot P(x_{(j)})\right]
\\
&= \argmax_{\theta}  \sum_{j=1}^{n \cdot s} \log \left[\frac{P(\tilde{z}_{(j)}, x_{(j)} | \theta) \cdot P(x_{(j)})}{P(x_{(j)} | \theta)} \right] \\
&\text{Independence of } X \text{ and } \theta \\
&= \argmax_{\theta}  \sum_{j=1}^{n \cdot s} \log \left[P(\tilde{z}_{(j)}, x_{(j)} | \theta) \right] \\
&= \argmax_{\theta} \log \mathcal{L}(\theta | \tilde{Z}, X)
\end{aligned}
\end{equation}

From Equation \ref{eq:theta-update} we can see that the M-step amounts to maximising the log-likelihood of the parameter $\theta$ with respect to the sampled values $\tilde{Z}$ and the given realisations of the random vector $X$. This in return means that, in the E-step, it suffices to produce the samples $\tilde{Z}$ that are needed in the M-step.

Since all elements $z_{k,l} \in \mathcal{Z}$ are independent of each other given $X$, the parameters $\theta$ can be separated between each distribution $z_{k,l} | X$ and are called $\theta_{k,l}$. With Equation \ref{eq:theta-update} it follows that

\begin{equation} \label{eq:theta-kl-update}
\small
\begin{aligned}
\theta_{k,l}^{(t+1)} &= \argmax_{\theta} \log \mathcal{L}(\theta | \tilde{z}_{k,l}, X)
\end{aligned}
\end{equation}

\subsection{SSM Explanation Templates}
\label{app:ssm-explanation-templates}

\begin{table}[h]
  \small
  \centering
  \begin{tabular}{|c|c|}
    \hline
    \textbf{Conditions} & \textbf{Template} \\
    \hline
    $Y = \text{contradiction}$ & "$x_{1,k}$ is not the same as $x_{2,l}$" \\
    $z_{k,l} = \text{contradiction}$ & \\
    $k = l$ & \\
    \hline
    $Y = \text{contradiction}$ & "If the $\text{subphrase}_k$ of sentence 1 is \\
    $z_{k,l} = \text{contradiction}$ &$x_{1,k}$, then the $\text{subphrase}_l$ \\
    $k \neq l$ & of sentence 2 cannot be $x_{2,l}$" \\
    \hline
    $Y = \text{entailment}$ & "$x_{1,k}$ is the same as $x_{2,l}$" \\
    $z_{k,l} = \text{entailment}$ & \\
    $k = l$ & \\
    \hline
    $Y = \text{entailment}$ & "If the $\text{subphrase}_k$ of sentence 1 is \\
    $z_{k,l} = \text{entailment}$ & $x_{1,k}$, then the $\text{subphrase}_l$ \\
     $k \neq l$ & of sentence 2 has to be $x_{2,l}$" \\
    \hline
    $Y = \text{neutral}$ & "There is no indication that the \\
    $z_{k,l} \neq \text{entailment} \ \forall k$ & $\text{subphrase}_l$ of sentence 2 is $x_{2,l}$" \\
    \hline
  \end{tabular}
  \caption{Templates for the explanations of the SSM predictions. $\text{subphrase}_i$ is "Subject" for $i=1$, "Verb" for $i=2$, "Object" for $i=3$, "Location" for $i=4$, and "Clothing" for $i=5$.}e
  \label{tab:templates}
\end{table}

\subsection{Instructions for Evaluating free-text Explanations}
\label{sec:human-instructions}

\subsection*{Background}




\subsection*{1. Evaluation of GPT-3.5 Explanations}

Please answer the following questions for each of the provided explanations produced by GPT-3.5. If not stated otherwise, mark the respective answer column in the evaluation template with 1 if the question is answered with yes.

\begin{enumerate}
    \item \textbf{Subphrases}: If any, which subphrases are mentioned in the explanation? Possible subphrases are subject, verb, object, clothing, and location either of sentence one or sentence two. Mark the answer column for the respective subphrase with 1 if the subphrase is mentioned in the explanation.
    \item \textbf{Structure}: Does the explanation describe the relationship between subphrases? This question should also be answered with yes if the explanation makes a statement about the relationship between subphrases that are not mentioned in question 1.
    \item \textbf{Support}: Does the explanation justify the predicted label? i.e.\ given that the explanation is true, is the predicted label correct?
    \item \textbf{Correctness}: Is the explanation factually correct? i.e.\ given your experience about our world and given the statements in sentence one and sentence two, is the explanation true?
\end{enumerate}

\subsection*{2. Evaluation of SSM Explanations}

The are explanations and predictions provided for two different versions of the statistical surrogate model (SSM). The first SSM's predicted labels and explanations are marked with subscript "large" and the predicted labels and explanations produced by the second SSM are marked with subscript "small". Please answer the following question for each of the provided explanations produced by the large and by the small SSM.

\begin{enumerate}
    \item \textbf{Overall Correctness}: Is the explanation factually correct given the premise and hypothesis? i.e.\ given your experience about our world and given the statements in sentence one and sentence two, is the explanation true?
    \item \textbf{Subphrase Correctness}: Is the explanation factually correct given the subphrases? i.e.\ given your experience about our world and given the subphrases extracted from sentence one and sentence two, is the explanation true?
\end{enumerate}

\subsection{Examples for SSM Output}
\label{app:example-outputs}

\begin{enumerate}
    \item \textbf{Premise:} A young woman sits crosslegged beside her purse on the grass among a crowd of dogs.\\
    \textbf{Hypothesis:} The woman is on the couch with the dogs.\\
    \textbf{NLE Small SSM:} Grass is not the same as couch.\\
    \textbf{NLE Large SSM:} Grass is not the same as couch.
    
    \item \textbf{Premise:} Two men are shopping for orange juice.\\
    \textbf{Hypothesis:} Two men are getting breakfast\\
    \textbf{NLE Small SSM:} There is no indication that the verb of sentence 2 is getting.\\
    \textbf{NLE Large SSM:} There is no indication that the object of sentence 2 is breakfast.

    \item \textbf{Premise:} A man tries to get himself into shape on a treadmill.\\
    \textbf{Hypothesis:} A man exercising.\\
    \textbf{NLE Small SSM:} Man is the same as a man and get is the same as exercising and if the location of sentence 1 is treadmill, then the verb of sentence 2 has to be exercising.\\
    \textbf{NLE Large SSM:} There is no indication that the subject of sentence 2 is a man and there is no indication that the verb of sentence 2 is exercising.
\end{enumerate}

\end{document}